# Chatbots put to the test in math and logic problems: A preliminary comparison and assessment of ChatGPT-3.5, ChatGPT-4, and Google Bard

Vagelis Plevris[1*], George Papazafeiropoulos[2] and Alejandro Jiménez Rios[3]

[1] Department of Civil and Architectural Engineering, Qatar University
P.O. Box: 2713, Doha, Qatar
e-mail: vplevris@qu.edu.qa

[2] School of Civil Engineering, National Technical University of Athens
Zografou Campus, 15780 Athens, Greece
e-mail: gpapazafeiropoulos@yahoo.gr

[3] Department of Built Environment, Oslo Metropolitan University
St. Olavs Plass, 0130 Oslo, Norway
e-mail: alejand@oslomet.no

*Corresponding author

**Abstract.** A comparison between three chatbots which are based on large language models, namely ChatGPT-3.5, ChatGPT-4 and Google Bard is presented, focusing on their ability to give correct answers to mathematics and logic problems. In particular, we check their ability to Understand the problem at hand; Apply appropriate algorithms or methods for its solution; and Generate a coherent response and a correct answer. We use 30 questions that are clear, without any ambiguities, fully described with plain text only, and have a unique, well defined correct answer. The questions are divided into two sets of 15 each. The questions of Set A are 15 "Original" problems that cannot be found online, while Set B contains 15 "Published" problems that one can find online, usually with their solution. Each question is posed three times to each chatbot. The answers are recorded and discussed, highlighting their strengths and weaknesses. It has been found that for straightforward arithmetic, algebraic expressions, or basic logic puzzles, chatbots may provide accurate solutions, although not in every attempt. However, for more complex mathematical problems or advanced logic tasks, their answers, although written in a usually "convincing" way, may not be reliable. Consistency is also an issue, as many times a chatbot will provide conflicting answers when given the same question more than once. A comparative quantitative evaluation of the three chatbots is made through scoring their final answers based on correctness. It was found that ChatGPT-4 outperforms ChatGPT-3.5 in both sets of questions. Bard comes third in the original questions of Set A, behind the other two chatbots, while it has the best performance (first place) in the published questions of Set B. This is probably because Bard has direct access to the internet, in contrast to ChatGPT chatbots which do not have any communication with the outside world.

## 1 Introduction and purpose of the study

Chatbots appeared in the 1960's and the first one was used to simulate the conversation between a psychotherapist and his/her patient [1]. A chatbot is a computer program or artificial intelligence (AI) system designed to engage in conversation with human users through text or voice-based interactions [2]. Chatbots can be used for various purposes, such as providing customer support, answering questions, offering recommendations, or even casual conversation [3]. They are typically designed to mimic human-like conversation patterns and can be found on messaging platforms, websites, and mobile applications. They are AI-driven programs designed to engage in natural language conversations with users. They often use natural language processing (NLP) and machine learning (ML) algorithms to understand and respond to user inputs more effectively.

Lately, in November 2022, the field of chatbots was revolutionized with the release of ChatGPT by OpenAI, due to its ability to generate more human-like and contextually coherent responses compared to previous models. The chatbot garnered attention for its detailed responses and articulate answers across many domains of knowledge and achieved a world record of reaching one million users within only 5 days since launch. In January 2023, it reached over 100 million users, making it the fastest growing consumer application to date [4].

Chatbots can be put to the test in various fields to evaluate their understanding and problem-solving abilities, and even be tested against actual professional examinations [5]. In the present study, we put three Chatbots to the test in math and logic problems, in particular (i) ChatGPT-3.5, (ii) ChatGPT-4, and (iii) Google Bard (or simply Bard), to determine their capacity to:

1. **Understand the problem**: Chatbots must be able to accurately interpret the user's input, recognize the type of problem being posed, and identify the relevant information needed to solve it.

2. **Apply appropriate algorithms or methods**: Chatbots need to utilize appropriate problem-solving techniques or mathematical algorithms to tackle the given problem. This may involve arithmetic, algebra, calculus, or logical reasoning, depending on the complexity of the problem.

3. **Generate a coherent response and the correct answer**: Once a solution is derived, chatbots should present the answer in a clear and concise manner, making sure the response is relevant and easy for the user to understand. Also, the final answer to the question should be mathematically correct.

Various investigations and surveys have appeared in the literature, only a few months after the release of ChatGPT and Bard chatbots, testing their mathematical capabilities and the degree up to which they could provide realistic help to professional mathematicians [6]. The skills of ChatGPT were tested across various logical reasoning domains, ranging from the ability to answer computational questions (e.g. calculate a complicated integral) or complete mathematical proofs that have gaps or missing steps, to the ability of even solving mathematical problems taken from Olympiads and across-domain thinking (for example asking the chatbot which theorems are required to prove a given theorem) [6]. The testing of ChatGPT was performed against a new dataset that was released in the same study [6]. The results showed that ChatGPT's mathematical abilities are significantly below those of an average mathematics graduate student. In another study it has been found that ChatGPT's performance changes dramatically if it is asked to provide explanations or extra text for its answer, compared to the case that it is asked just to give the answer only with absolutely no other text [7]. In the former case its probability of failure was estimated as 20% whereas in the latter case it was 84%. In this study, the DRAW-1K dataset [8-10] was used to evaluate the performance of ChatGPT. An additional attempt to address and resolve the inability of ChatGPT to solve complex mathematical problems has been made in [11], according to which the inability of ChatGPT to provide faithful answers to (mathematical) questions is addressed by categorizing ChatGPT's failures into four types (comprehension, factuality, specificity, and inference) and establishing three critical abilities associated with quality assurance failures (knowledge memorization, knowledge association and knowledge reasoning). It was concluded in [11] that furnishing the model with fine-grained external knowledge, hints for knowledge association, and guidance for reasoning can empower the model to answer questions more faithfully.

Despite the continuously growing work on investigation and testing ChatGPT, to the best of the authors' knowledge, analogous studies have not been published for testing other major chatbots, such as Bard for example, or even more comparing different chatbots in terms of the correctness of their answers. The present study tries to make such a performance comparison of different chatbots based on the answers they provide in a series of well-posed logical and mathematical questions that have been posed to them. Testing the chatbots in such problems can help researchers, developers, and users assess their capabilities and limitations. These tests can reveal areas where chatbots excel, as well as areas that require improvement. Ultimately, these evaluations contribute to the ongoing development of more advanced, capable, and user-friendly AI chatbot systems.

The remainder of the paper is organized as follows: Section 2 discusses the three chatbots used in the study, the technologies behind them and the differences among the three models. Section 3 presents the methodology of the study, which is followed by the discussion of the individual answers to each question, in Section 4. Section 5 presents and discusses the performance of the three chatbots in the 30 questions. Finally, section 6 presents the conclusions of this work and some future directions.

## 2 Chatbots used in the study

In this study we have used three chatbots: (i) ChatGPT-3.5, (ii) ChatGPT-4 and (iii) Google Bard. All these chatbots rely on a large language model (LLM), which is a type of AI model designed to understand and generate human-like text by leveraging deep learning (DL) techniques. These models are "large" in the sense that they have a massive number of parameters, often ranging from hundreds of millions to hundreds of billions, which allows them to capture complex language patterns and relationships.

### 2.1 ChatGPT-3.5 and ChatGPT-4

GPT, short for Generative Pre-trained Transformer, is a state-of-the-art language model developed by OpenAI. It comes in two main versions today, GPT-3.5 and GPT-4. On the other hand, ChatGPT is a chatbot app, powered by GPT, which is optimized for dialogue. Consequently, ChatGPT refers to a chatbot which is powered by GPT (any version), ChatGPT-3.5 refers to the chatbot which is powered by GPT-3.5 and similarly chatGPT-4 refers to the chatbot which is powered by GPT-4.

ChatGPT is designed to generate human-like text responses based on the input it receives. At a high level, ChatGPT works by using DL techniques to understand and generate text. It is trained on a massive amount of data from the internet, which allows it to learn the patterns and structures of language. During training, the model predicts the next word in a sentence given the previous words, and this process is repeated many times to learn the relationships between words and the overall context of the text. When a user interacts with ChatGPT, they provide it with a prompt or a question. The model then analyzes the input and generates a response based on its understanding of the given text and the patterns it has learned. The response is generated by sampling from a probability distribution over the vocabulary of possible words, considering the context and the likelihood of each word in the given context.

The training process for ChatGPT involves training on a vast amount of text data, but it is important to note that the model does not have real-time access to the internet or current events. The knowledge and information available to ChatGPT are based on the data it was trained on, which has a cutoff date in September 2021. As a result, the model may not be aware of recent developments or be able to provide real-time information.

In scientific terms, ChatGPT operates through the following key components and processes:

1. **Transformer architecture**: The Transformer model, proposed by Vaswani et al. in 2017 [12], is the backbone of ChatGPT. It uses self-attention mechanisms to process input data in parallel rather than sequentially, allowing it to efficiently handle long-range dependencies in text.

2. **Pre-training**: ChatGPT is pre-trained on a large corpus of text from the internet. During this unsupervised learning phase, it learns the structure and statistical properties of the language, including grammar, vocabulary, and common phrases. The objective is to predict the next word in a sentence, given the context of the preceding words.

3. **Fine-tuning**: After the pre-training phase, ChatGPT is fine-tuned on a smaller dataset, often containing specific conversational data. This supervised learning phase involves training the model to generate appropriate responses in a conversational setting. The model learns from human-generated input-output pairs, refining its ability to provide contextually relevant and coherent responses.

4. **Tokenization**: When ChatGPT receives input text, it tokenizes the text into smaller units, such as words or subwords. These tokens are then mapped to unique IDs, which are used as input for the model.

5. **Encoding and decoding**: The Transformer architecture consists of an encoder and a decoder. The encoder processes the input tokens, while the decoder generates the output tokens sequentially. Both the encoder and decoder rely on self-attention mechanisms and feed-forward neural networks to process and generate text.

6. **Attention mechanisms**: Attention mechanisms enable the model to weigh the importance of different parts of the input when generating a response. This helps ChatGPT focus on the most relevant information in the input text and generate coherent and contextually appropriate responses.

7. **Probability distribution**: The model's output is a probability distribution over the vocabulary for the next token in the sequence. The token with the highest probability is chosen, and this process is repeated until the model generates a complete response or reaches a predefined maximum length.

8. **Beam search or other decoding strategies**: To generate the most likely response, ChatGPT uses decoding strategies like beam search, which maintains a set of top-$k$ candidate sequences at each time step. These strategies help in finding a balance between fluency and coherence while minimizing the risk of generating nonsensical or overly verbose outputs.

By combining these components and techniques, ChatGPT can understand and generate human-like text, making it a powerful tool for various applications, such as conversational agents, content generation, and question-answering systems. ChatGPT can work with various languages, to some extent. While it is primarily trained on English text data, it also learns from multilingual text sources during its pre-training phase. As a result, it can understand and generate text in several languages, such as Spanish, French, German, Chinese, and more. However, it is important to note that ChatGPT's proficiency in different languages may vary depending on the amount and quality of training data available for each language. Its performance is typically better for languages with a larger presence on the internet and in the training data. Naturally, it exhibits best performance in the English language.

It is worth mentioning that while ChatGPT is impressive in its ability to generate coherent responses, it can sometimes produce incorrect or nonsensical answers. It is essentially a statistical model that relies on patterns in the data, and it probably does not possess true understanding or knowledge like a human. Therefore, it is important to critically evaluate and fact-check the information provided by ChatGPT. OpenAI continues to refine and improve language models like ChatGPT, and future iterations may address some of the limitations and challenges associated with the current generation models.

### GPT-3.5 vs GPT-4

GPT-4 is an advanced version of its predecessor, GPT-3.5. It was released on March 14, 2023. Some of the improvements and advanced features of GPT-4 include:

1. **Increased model size**: GPT-3 has 175 billion parameters that allow it to take an input and give a text output that best matches the user request. GPT-4 has way more parameters, although the exact number is not known, leading to improved language understanding and generation capabilities. OpenAI has not given information about the size of the GPT-4 model [13].

2. **Enhanced context understanding**: GPT-4 can handle longer text inputs and maintain context better, which allows for more coherent and relevant responses.

3. **Improved fine-tuning**: GPT-4 has been fine-tuned on more diverse and specific tasks, allowing it to perform better across a wider range of applications.

4. **Better handling of ambiguity**: GPT-4 is better at resolving ambiguous input and providing clearer, more accurate responses.

5. **More robust language support**: GPT-4 has been trained on a broader range of languages and can better handle multilingual tasks and code-switching.

6. **Enhanced safety and ethical considerations**: GPT-4 has been designed with more robust safety measures to prevent harmful outputs, ensuring better alignment with human values.

7. **Domain-specific knowledge**: GPT-4 has been trained on more specific knowledge domains, allowing it to provide more accurate information and support specialized tasks.

These are general features and improvements over the previous version. The actual performance of GPT-4 may still vary depending on the specific task, input, or context. ChatGPT-3.5 is free for all users and does not have any limitations in its use, while ChatGPT-4 currently has a cap of 25 messages every 3 hours and it is provided as a paid service (ChatGPT Plus), with a $20/month subscription.

## 2.2 Google Bard

Google Bard is a LLM chatbot developed by Google AI. It is trained on a massive dataset of text and code, including Wikipedia, Books, Code, Stack Overflow, Google Search, and other publicly available datasets. It can generate text, translate languages, write different kinds of creative content, and answer user questions in an informative way. Bard was released for US and UK users on March 21, 2023, and it is free to use. According to Google, it is powered by several technologies, including:

- **Natural language processing (NLP)**: NLP is a field of computer science that deals with the interaction between computers and human (natural) languages. NLP is used in Bard to understand and process the text that the user inputs.

- **Machine learning (ML)**: ML is a field of computer science that gives computers the ability to learn without being explicitly programmed. ML is used in Bard to train the model on the massive dataset of text and code.

- **Deep learning (DL)**: DL is a subset of ML that uses artificial neural networks to learn from data. DL is used in Bard to train the model to generate text, translate languages, write different kinds of creative content, and answer user questions in an informative way.

As of May 2023, Bard is still officially under development, but it has learned to perform many kinds of tasks, such as: (i) Follow user instructions and complete user requests thoughtfully; (ii) Use its knowledge to answer questions in a comprehensive and informative way, even if they are open ended, challenging, or strange; and (iii) Generate different creative text formats of text content, like poems, code, scripts, musical pieces, email, letters, etc. Bard has direct access to the internet and its data is constantly being updated, so it is always learning new things.

## 2.3 Differences between ChatGPT and Bard

A key difference between ChatGPT and Bard is that Bard has access to Google search, and it continually draws data from the internet, so it has the latest information. On the other hand, ChatGPT does not have direct access to the internet or any specific data source. Its knowledge is based on the vast amount of text data that it was trained on, which includes web pages, books, articles, and other textual sources. The training data of ChatGPT was last updated in 2021. As a result, ChatGPT does not have knowledge of the latest events and developments in any scientific or other field.

## 3 Methodology of the study

In this study, we use 30 questions describing mathematics and logic problems that have a unique correct answer. These questions are fully described with plain text only, without the need for any images or special text formatting. The questions are divided into two sets of 15 questions each (Set A and Set B). The questions of Set A are 15 "Original" problems that cannot be found online, at least in their exact wording, while Set B contains 15 "Published" problems that one can find online by searching on the internet, usually with their solution. Each question is posed three times to each chatbot. For ChatGPT-3.5 and ChatGPT-4, we had to click the button "Regenerate response" two times to receive three answers. With Bards things were simpler as it automatically provides three "draft" answers. The first answer is displayed to the user, while to see the other two, one needs to click the "View other drafts" button. The full set of the 30 questions, together with the correct answer for each one of them, an explanation of the solution and the $30 \times 3 \times 3 = 270$ detailed answers of the chatbots can be found in the relevant published dataset [14].

A hypothesis that needs to be tested is that Chatbots that rely primarily on online "ready-to-use" information and online search may be better in answering the questions of Set B, but they will have problems with those of Set A. It must be noted that we tried to avoid ambiguities and to make the problems as clear as possible. Therefore, we do not try to check the abilities of the chatbots in handling ambiguous or ill-posed problems. We also do not engage in any kind of dialogue with the chatbot for any question and we do not allow it to ask clarifying questions. To keep things simple and fair, we do not provide any feedback to any answer given by a chatbot. Thus, we do not check the ability of a chatbot to learn from user feedback. In general, chatbots should be able to learn from the user's feedback and improve their problem-solving abilities over time, but this is not an aim of the present study. The focus of the study is on checking the ability of chatbots to (i) Understand the problem at hand, (ii) Apply appropriate algorithms or methods, and (iii) Generate a coherent response and the correct answer.

## 4    Discussion of the individual answers to each question

It this section we provide each full question, its correct answer, the score of each chatbot and some discussion on the responses given. Each question was posed three times and the chatbots gave three answers for each question. A score of *k-l-m* means *k* correct answers for ChatGPT-3.5, *l* correct answers for ChatGPT-4 and *m* correct answers for Bard, in their three attempts. We keep this order throughout the manuscript.

### 4.1 Set A: "Original" questions

This is a set of 15 "Original" questions, denoted as A01 to A15, which cannot be found online and have not been published previously, at least with the same wording.

Questions A01 and A02 (Scores: 3-1-1 and 0-0-0)

*A01. "Solve the following cubic equation: $x^3 - 13*x^2 + 50*x - 56 = 0$"*

*A02. "Solve the following cubic equation: $100*x^3 - 1340*x^2 + 5389*x - 6660 = 0$"*

The correct answer to question A01 is "**x=2, x=4, x=7**". The correct answer to question A02 is "**x=2.5, x=3.7, x=7.2**". Both questions are of similar nature; finding the numerical values of *x* that will satisfy the given cubic equation. All three chatbots seem to understand the nature of the problem. In terms of the methods or algorithms used to solve the problem, ChatGPT-3.5 implements the rational roots theorem in 5 out of 6 times, and the Cardano's formula once. ChatGPT-4 attempts to provide a solution by using the rational roots theorem, a graphical solution, and a code snippet in python in 66.7%, 16.7% and 16.7% of the times, respectively. Finally, Bard uses factor lists 5 times and the rational roots theorem once. All the implemented methods or algorithms can correctly lead to a right answer; thus, it could be said that the chatbots have chosen a proper way to give an answer.

The first problem (A01) has three integer roots. ChatGPT-3.5 had the best performance in this relatively easy task. ChatGPT-3.5 managed to give 3 correct solutions, while ChatGPT-4 failed in the first attempt, got it right in the second and failed again in the third. Bard found it correctly on the first attempt but missed the other two. It is impressive that although these models are so complex and can find solutions to difficult problems, they can fail in such an easy task. In addition, it must be highlighted that normal reasoning appears not to work for them. After solving this exercise, a human would easily check the solution by substituting the found values for *x* in the equation to see if the equation is satisfied. This is not the case with AI chatbots. They did not bother to check their final solution.

Problem A02 has three roots which are not integers, so this task is a bit harder than the previous one. All models fail to give a correct solution, in all their attempts in this question.

Question A03 (Score: 1-1-0)

*"A closed club of professional engineers has 500 members. Some members are "old members" while the others are "new members" (subscribed within one year from now). An event was organized where old members had to pay $200 each for their participation while new members had to pay $140 each. The event was successful and while all new members came, only 70% of the old members attended. What is the amount of money (in $) that was collected from all members for this event?"*

The correct answer is "**70,000**". This is a relatively "wordy" question where the chatbots need to understand not only the numbers given, but also the relationship created with the words between those numbers. The question is only asked at the end. As all chatbots try to provide an amount of money as an answer, it could be said that they correctly understood the question. Although all three attempt to provide a solution doing some mathematical calculation, it seems that only ChatGPT chatbots have the capacity of correctly defining "unknowns" and assigning given data to variables, which could be considered as the "correct" methodology. On the other hand, Bard seems to work with the numbers given without assigning these numbers an actual problem-context meaning.

This question was very challenging for all chatbots. Only ChatGPT-3.5 and ChatGPT-4 got it correct once. Bard failed in all attempts. The challenge with this question is that we do not really know the number of the old members and the number of new members. Nevertheless, these are not asked by the problem. The question simply asks the amount of money collected, which does not require knowing the exact number of old and new members. This was the tricky part of this question, which caused problems to the chatbots.

Question A04 (Score: 1-3-0)

*"The sum of three adults' ages is 60. The oldest of them is 6 years older than the youngest. What is the age of each one of them? Assume that an adult is at least 18 years old."*

The correct answer is "**18, 18, 24**". ChatGPT-3.5 gets it correct one time again, ChatGPT-4 gets it correct all three times, as it clearly understands the problem. Bard still fails in all the attempts it made. This is a straightforward mathematical problem but with some constraints included. All three chatbots attempt to compute the age of the three adults, which somehow shows their understanding of the question asked. Furthermore, they try to use algebra and can assign numerical values to variables (a quality which Bard did not show in the previous question A03) to come up with a solution, implementing a "proper" methodology.

Question A05 (Score: 3-3-1)

*"A decade ago, the population of a city was 55,182 people. Now, it is 170% larger. What is the city's current population?"*

The correct answer is "**148,991 people**". Interestingly, only ChatGPT seem to fully understand the problem, whereas Bard did not even correctly identify the city's current population in 2 out of 3 attempts. On the other hand, the three seem to implement appropriate methods or algorithms to give an answer to the question. Both ChatGPT models get this right, in all their attempts. Strangely, Bard fails 2 out of 3 times, although the problem seems clear and simple. In these two attempts, Bard fails to understand that 170% larger means 270% of the original value (not 170% of it).

Question A06 (Score: 1-3-0)

*"What is the precise sum of 523,654,123 and 7,652,432,852,136?"*

The correct answer is "**7,652,956,506,259**". In this case all three chatbots seem to understand the problem, which is a simple mathematical addition. Strangely, ChatGPT-3.5 fails to provide a correct result on 2 out of 3 attempts and Bard fails in all three attempts. GPT-4 gets it correct all three times. It is weird that two chatbots fail in this rather simple calculation. Table 1 shows the responses of ChatGPT-3.5 and Bard in this question. These models fail to predict one or two digits of the result, while the other digits are correct, which is also weird.

**Table 1**. Responses of ChatGPT-3.5 and Bard in the addition question A06.

| | |
|---|---|
| **Correct calculation and result** | 523,654,123<br>+7,652,432,852,136<br>**7,652,956,506,259** |
| **ChatGPT-3.5 Attempt #1** | ✓ 7,652,956,506,259 |
| **ChatGPT-3.5 Attempt #2** | ✗ 7,653,956,506,259 |
| **ChatGPT-3.5 Attempt #3** | ✗ 7,653,956,506,259 |
| **Bard Attempt #1** | ✗ 7,652,956,515,259 |
| **Bard Attempt #2** | ✗ 7,652,956,505,259 |
| **Bard Attempt #3** | ✗ 7,652,956,505,259 |

Question A07 (Score: 2-3-3)

*"You decide to make a road-trip with your new car. The distance between City A and City B is 120 km. When you travel from A to B, your average speed is slow, 60 km/h. When you travel from B to A, your average speed is high, 120 km/h. What is the average speed for the whole trip A to B to A (with return to City A)?"*

The correct answer is "**80 km/h**". Many people can get confused thinking that the answer is 90 km/h which is the average of 60 km/h and 120 km/h, but this is not a correct approach. This question presents one of the best chatbots performances so far (88.9% correct responses). All three chatbots seemed to understand the problem correctly and to implement appropriate methods/algorithms (algebra-based) to come up with an answer. Chatbots succeed in providing the correct answer in all their attempts, except for ChatGPT-3.5 which failed one time. Similar questions can be found online, but not with the exact wording of this problem.

Question A08 (Score: 1-3-3)

*"If Tom has 35 marbles and I have 12 marbles, and then Tom gives me 9 marbles, how many more marbles does Tom have than I?"*

The correct answer is "**5**". After a careful examination of the performance of ChatGPT-3.5, it could be observed that it systematically fails in 2 out of 3 attempts. The reason for this is that the chatbot fails to "understand" that in this context "giving" also means "losing" (although this issue was not present in the last attempt of the chatbot). Although this appears to be a very easy question, ChatGPT-3.5 failed in two of its attempts, while the other two Chatbots got it correct all three times.

Question A09 (Score: 3-3-3)

*"Tom's father has three children. The younger child's name is Erica. The middle child's name is Sam. What is the name of the older child?"*

The correct answer is "**Tom**". All chatbots managed to correctly interpret the problem, implement a suitable method/algorithm to find a solution, and give a correct answer to this simple question, in all their three attempts, thus showing a remarkable high performance while dealing with purely logical problems.

Question A10 (Score: 3-3-1)

*"A woodworker normally makes a certain number of parts in 11 days. He was able to increase his productivity by 3 parts per day, and so he not only finished the job 2 days earlier, but in addition he made 9 extra parts. How many parts does the woodworker normally make per day?"*

The correct answer is "**9**". Both versions of ChatGPT manage to give correct answers to this question, in all their three attempts. Bard was correct only in its first attempt and failed the other two attempts. In this occasion, all chatbots seem to understand the problematic, which is related to the computation of number of parts the woodworker produced per day, originally. Moreover, all three use variables and apply basic algebra operations to find the solution, which could indeed be considered as an appropriate methodology.

Question A11 (Score: 2-3-0)

*"Think of a number. Add 5, double the result, then subtract 12, then take half of the result and finally subtract the initial number. What is the result?"*

Here the correct answer is "**-1**". In this question, ChatGPT-3.5 got 2/3, GPT-4 got them all correct, while Bard failed three times. This is a simple, yet interesting logico-mathematical problem that requires to follow a process step by step and apply basic operations (+, -. *, /) in each one to find the correct solution. All three chatbots seem to correctly understand the problematic. Furthermore, both ChatGPT chatbots applied a correct multi-step algorithm to find a solution. On the other hand, Bard tried to pose the problem as an equation, sometimes even generating a greater number of variables than required.

Question A12 (Score: 3-2-3)

*"If one and a half hens lay one and a half eggs in one and a half days, how many eggs do 9 hens lay in 9 days?"*

The correct answer is "**54**". In this question, ChatGPT-3.5 and Bard got it correct 3 times, while GPT-4 failed once and got it correct two times. This is a relatively simple question in mathematical terms, but it is posed in a tricky way in terms of language, as the "*one and a half*" string is repeated several times, which may pose a challenge to a chatbot. Although all three chatbots seem to correctly "understand" the problematic and apply an adequate methodology to solve it, on one occasion, ChatGPT-4 seems to get "confused" in the process. It adequately determines that 1 hen lays 1 egg per day, but then it assigns the original number mentioned in the question (1.5) and the result turns out to be erroneous. Variations of this problem can be found online, but not with this exact wording.

Question A13 (Score: 0-2-0)

*"Find a 4-digit number so that the last four digits of the number squared is the number itself."*

The correct answer is "**9376**". Variations of this problem can be found online, but not with this exact wording. In this question, ChatGPT-3.5 and Bard fail three times, while GPT-4 gets it correct two times. The one time it fails, it gives 0625 as an answer stating that "Note that 0625 is technically a 4-digit number, although it may appear as a 3-digit number (625) in some representations due to the leading zero." This shows that GPT-4 has understood the problem and that the solution it provided might not be what was expected. It tries to defend its answer with some reasoning, which is interesting. It must be noted that in its second attempt, ChatGPT-4 states that "such a number is called a "Kaprekar number"", which is not correct as a definition. Nevertheless, ChatGPT-4 ends up with the correct answer in the end.

<u>Question A14 (Score: 0-3-0)</u>

*"The number of water lilies on a lake doubles every two days. If there is initially one water lily on the lake, it takes exactly 50 days for the lake to be fully covered with water lilies. In how many days will the lake be fully covered with water lilies, if initially there were two water lilies (identical with the previous one) on it?"*

The correct answer is "**48**". Variations of this problem can be found online, but not with this exact wording. In most similar problems, it is stated that the number doubles every day. Both ChatGPT-3.5 and Bard failed three times in this problem. GPT-4 got it correct three times. Although by the answer given by all three chatbots it could be said that all of them correctly understood the question (they all "acknowledge" the fact that the number of lilies doubles every second day), only ChatGPT-4 seems to apply the correct "logic" or algorithm to come up with correct answers. It is also interesting to note how Bard may be completely biased in this specific problem due to the large number of online appearances of this type of problem when it is stated that the number of lilies would double every day. Under such a scenario, the correct answer would be 49, which is the answer given by Bard in all three attempts, in other words, it seems to give the "right" answer to the "wrong" question.

<u>Question A15 (Score: 1-3-3)</u>

*"There are 25 handball teams playing in a knockout competition (i.e. if you lose a match, you are eliminated and do not continue further). What is the minimum number of matches (in total) they need to play to decide the winner?"*

The correct answer is "**24**". Variations of this problem can be found online, but not with this exact wording. The concept of such a tournament is a concept that can be found broadly on the internet. ChatGPT-4 and Bard get it correct three times, while ChatGPT-3.5 fails two times.

## 4.2 Set B: "Published" questions

This is a set of 15 "Published" questions (problems), denoted as B01 to B15, which can be found online. The first 9 questions were taken from [15]. Questions 10 and 15 are from [16], while questions 11 and 12 are from [17]. Finally, question 13 is from [18] and question 14 is taken from [19].

<u>Question B01 [15] (Score: 2-3-2)</u>

*"A bad guy is playing Russian roulette with a six-shooter revolver. He puts in one bullet, spins the chambers and fires at you, but no bullet comes out. He gives you the choice of whether or not he should spin the chambers again before firing a second time. Should he spin again?"*

The correct answer is "**Yes**". ChatGPT-3.5 got it right 2 times in three attempts, GPT-4 got it right in all three attempts, while Bard missed the first but got it right in the other two. From a purely mathematic point of view, this is a simple problem of probabilities, i.e., whether the outcome is desirable or not would determine if a higher or a lower probability percentage is selected as answer. Nevertheless, it could be argued that the wording used turns the question into an ethical one, by describing a person as "bad" and the corresponding assumptions that would imply. This could explain the fact that not all answers were correct as this ethical dilemma may have confused the chatbots.

Question B02 [15] (Score: 2-3-3)

*"Five people were eating apples, A finished before B, but behind C. D finished before E, but behind B. What was the finishing order?"*

The correct answer is "**CABDE**". This is an easy problem. Both GPT-4 and Bard give correct answers in all their attempts, while ChatGPT-3.5 misses it one time. All three chatbots seem to "understand" the problematic correctly as they all attempted to put the given letters in order. Furthermore, the method applied by the chatbots also seem correct as the correct order of the letters can be known based on the information provided by the different statements of the question, which is the information used by the chatbots to attempt providing a right answer.

Question B03 [15] (Score: 0-0-1)

*"A man has 53 socks in his drawer: 21 identical blue, 15 identical black and 17 identical red. The lights are out, and he is completely in the dark. How many socks must he take out to make 100 percent certain he has at least one pair of black socks?"*

The correct answer is "**40**". This problem is not so easy to solve, although once given the solution everybody can understand that it makes sense. In this question both GPT models fail in all their attempts. Strangely, ChatGPT-3.5 gives answers which are more reasonable and closer to being correct. Bard gets it right once, in its first attempt. Regardless of the low performance of the chatbots (overall success of only 11.1%), they all seem to have some understanding of the problem. Both ChatGPT-3.5 and Bard seem to apply an adequate algorithm, which consists of discarding all socks of different color to the one of interest (although they failed on computing the right number), whereas ChatGPT-4 applies the wrong logic to find the answer.

Question B04 [15] (Score: 1-1-3)

*"Susan and Lisa decided to play tennis against each other. They bet $1 on each game they played. Susan won three bets and Lisa won $5. How many games did they play?"*

The correct answer is "**11**". This is a relatively easy problem which nevertheless caused trouble to the chatbots. Both ChatGPT-3.5 and GPT-4 failed two times and predicted the outcome correctly one time, while Bard gets it correct in all three attempts. The three chatbots appear to have correctly understood the situation. Both ChatGPT chatbots implemented a similar solving methodology based on solving algebraically a couple of equations (although not always successfully), whereas Bard provides the correct answer on three attempts apparently purely out of self-logic "reasoning", which may be a hint for the chatbot finding the solution of the problem posted online in the referenced source or elsewhere.

Question B05 [15] (Score: 3-3-3)

*"Jack is looking at Anne. Anne is looking at George. Jack is married, George is not, and we don't know if Anne is married. Is a married person looking at an unmarried person?"*

The correct answer is "**Yes**". This is a relatively easy problem which caused no trouble to the chatbots. All of them got it correct, three times. All chatbots "understood" correctly the problem and applied adequate methods/algorithms to solve it. Once again, the high performance of all three chatbots is ascertained while facing purely logic problems (as was the case with question A09).

Question B06 [15] (Score: 0-2-3)

*"A girl meets a lion and unicorn in the forest. The lion lies every Monday, Tuesday and Wednesday and the other days he speaks the truth. The unicorn lies on Thursdays, Fridays and Saturdays, and the other days of the week he speaks the truth. "Yesterday I was lying," the lion told the girl. "So was I," said the unicorn. What day is it?"*

The correct answer is "**Thursday**". In this question ChatGPT-3.5 fails three times. GPT-4 gets it right two times and fails once, while Bard gets it right in all its attempts. The three chatbots seem to "understand" the problematic in all attempts, they "know" that they must give a day of the week as an answer to the problem. Moreover, they use the statements of the problem to "reason" and come up with an answer, similarly to what a human would do. Once again this is a pure logic problem, but contrary to the cases of questions A09 and B05, the accuracy performance of the chatbots is considerably poorer.

Question B07 [15] (Score: 0-3-1)

*"Three men are lined up behind each other. The tallest man is in the back and can see the heads of the two in front of him; the middle man can see the one man in front of him; the man in front can't see anyone. They are blindfolded and hats are placed on their heads, picked from three black hats and two white hats. The extra two hats are hidden and the blindfolds removed. The tallest man is asked if he knows what color hat he's wearing; he doesn't. The middle man is asked if he knows; he doesn't. But the man in front, who can't see anyone, says he knows. How does he know, and what color hat is he wearing?"*

The correct answer is "**Black**". In this question, ChatGPT-3.5 fails three times and GPT-4 gets it right three times. Bard gets it correct only in the 3$^{rd}$ attempt. This is again a purely logical problem, although expressed in a relatively long text question, which may increase the level of difficulty for a LLM model. Nevertheless, all three chatbots seem to understand correctly the problematic and apply logic "reasoning" to come up with a solution. Nevertheless, their performance is far from satisfactory, except for ChatGPT-4.

Question B08 [15] (Score: 1-1-3)

*"A teacher writes six words on a board: "cat dog has max dim tag." She gives three students, Albert, Bernard and Cheryl each a piece of paper with one letter from one of the words. Then she asks, "Albert, do you know the word?" Albert immediately replies yes. She asks, "Bernard, do you know the word?" He thinks for a moment and replies yes. Then she asks Cheryl the same question. She thinks and then replies yes. What is the word?"*

The correct answer is "**Dog**". In this question, only Bard gets it correct three times, while the other two chatbots fail two times and get it correct only once. The complete reasoning of ChatGPT-3.5 in its second attempt does not appear to be 100% correct. Nevertheless, it comes up with the correct answer in the end, and for this reason we consider the answer as finally "Correct" in this case.

Question B09 [15] (Score: 0-0-3)

*"There are three people (Alex, Ben and Cody), one of whom is a knight, one a knave, and one a spy. The knight always tells the truth, the knave always lies, and the spy can either lie or tell the truth. Alex says: "Cody is a knave." Ben says: "Alex is a knight." Cody says: "I am the spy." Who is the knight, who the knave, and who the spy?"*

The correct answer is "**Alex: knight, Ben: spy, Cody: knave**". In this question, only Bard gets it correct three times, while the other two chatbots fail in all their three attempts. All three chatbots understood this is a word puzzle and implemented adequately a reasoning strategy based on the statements of the question to attempt providing the correct answer. Interestingly, none of the ChatGPT chatbots got any attempt correct, whereas Bard showed 100% accuracy. This may be because Bard was able to locate the right answer in the online source.

Question B10 [16] (Score: 0-0-3)

*"Kenny, Abby, and Ned got together for a round-robin pickleball tournament, where, as usual, the winner stays on after each game to play the person who sat out that game. At the end of their pickleball afternoon, Abby is exhausted, having played the last seven straight games. Kenny, who is less winded, tallies up the games played: Kenny played eight games. Abby played 12 games. Ned played 14 games. Who won the fourth game against whom?"*

The correct answer is "**Ned beat Kenny in the fourth game**". In this question, only Bard gets it correct three times, while the other two chatbots fail in all their three attempts.

Questions B11 [17] and B12 [17] (Scores: 3-3-2 and 2-3-2)

*B11. "The distance between two towns is 380 km. At the same moment, a passenger car and a truck start moving towards each other from different towns. They meet 4 hours later. If the car drives 5 km/hr faster than the truck, what are their speeds?"*

*B12"A biker covered half the distance between two towns in 2 hr 30 min. After that he increased his speed by 2 km/hr. He covered the second half of the distance in 2 hr 20 min. Find the distance between the two towns."*

The correct answer to B11 is "**Truck's speed: 45 km/hr, Car's speed: 50 km/hr**". In this relatively easy question, ChatGPT-3.5 and GPT-4 get is correct in all their attempts while Bard makes a mistake in its 2$^{nd}$ attempt and gets it correct in the other two. The problematic posed by the question is correctly identified by all three chatbots, which can be seen as they all try to give speeds as correct answers to the problem. Furthermore, they all implement a correct methodology to find a suitable solution, based on assigning the unknown speeds of the vehicles to variables and using algebraic operations to come up with the answer.

The correct answer to B12 is "**140 km**". In this easy question, only GPT-4 manages to give three correct answers. The other two models make one mistake and get it correct in the other two attempts.

Question B13 [18] (Score: 2-3-1)

*"Rhonda has 12 marbles more than Douglas. Douglas has 6 marbles more than Bertha. Rhonda has twice as many marbles as Bertha has. How many marbles does Douglas have?"*

The correct answer is "**24**". Although this appears to be a very easy question, only GPT-4 manages to give three correct answers. This is a relatively simple mathematical question where solution can be found through assigning variables to the unknowns and solving a relatively simple system of equations, which all three chatbots seem to understand in all their attempts. Nevertheless, ChatGPT-3.5 gets it correct two times and fails once while strangely Bard fails in two attempts and gets it correct only once.

Question B14 [19] (Score: 3-3-3)

*"15 workers are needed to build a wall in 12 days. How long would it take to 10 workers to build the same wall?"*

The correct answer is "**18 days**". This question posed a relatively easy mathematical problem where all chatbots managed to give correct answers, in all their three attempts. The perfect scoring reflects a good problematic understanding and adequate implementation of a method/algorithm to come up with an answer, for all three chatbots.

Question B15 [16] (Score: 0-0-3)

*"A hen and a half lay an egg and a half in a day and a half. How many eggs does one hen lay in one day?"*

The correct answer is "**⅔ of an egg**". Although this problem is similar to the previous one (B14), only Bard gets it correct three times in this case. The other two models strangely fail in all their attempts. This trend in the performance of the chatbots turns out to be quite interesting. It could be said that all three chatbots understood the problem and correctly attempted to provide the number of eggs as an answer. For the ChatGPT chatbots we are given the impression that the relatively "complex wording" of the question (the "and a half" string and the words "hen", "egg", and "day" are repeated several times in a quite short string) may have "confused" the chatbots, thus causing their failure to provide the correct answer. On the other hand, Bard's perfect score, may be again due to the fact that the answer can be easily found online. This question is similar to question A12 from Set A where the score was 3-2-3 for the three chatbots and Bard got it again correct three times. It is interesting that ChatGPT chatbots had a much better performance in question A12 in comparison to this question B15.

## 5    Discussion on the performance of chatbots

Due to space limitations, the detailed responses of the chatbots in all problems are not included in this paper, but they can be found in the published open access dataset [14]. Table 2 presents the scores of each chatbot in the first set of questions (Set A, 15 "Original" problems), for each question, and the relevant sums. Each chatbot gets 1 point for a correct answer and 0 points for an incorrect answer. In the table, correct answers are highlighted with green color, while incorrect ones are highlighted with red color.

**Table 2**. Responses of the three chatbots to the questions of Set A ("Original" problems) [*].

| Question | Chat GPT-3.5 | | | | GPT-4 | | | | Bard | | | |
|---|---|---|---|---|---|---|---|---|---|---|---|---|
| | #1 | #2 | #3 | SUM | #1 | #2 | #3 | SUM | #1 | #2 | #3 | SUM |
| A01 | 1 | 1 | 1 | 3 | 0 | 1 | 0 | 1 | 1 | 0 | 0 | 1 |
| A02 | 0 | 0 | 0 | 0 | 0 | 0 | 0 | 0 | 0 | 0 | 0 | 0 |
| A03 | 0 | 1 | 0 | 1 | 0 | 1 | 0 | 1 | 0 | 0 | 0 | 0 |
| A04 | 1 | 0 | 0 | 1 | 1 | 1 | 1 | 3 | 0 | 0 | 0 | 0 |
| A05 | 1 | 1 | 1 | 3 | 1 | 1 | 1 | 3 | 0 | 1 | 0 | 1 |
| A06 | 1 | 0 | 0 | 1 | 1 | 1 | 1 | 3 | 0 | 0 | 0 | 0 |
| A07 | 1 | 0 | 1 | 2 | 1 | 1 | 1 | 3 | 1 | 1 | 1 | 3 |
| A08 | 0 | 0 | 1 | 1 | 1 | 1 | 1 | 3 | 1 | 1 | 1 | 3 |
| A09 | 1 | 1 | 1 | 3 | 1 | 1 | 1 | 3 | 1 | 1 | 1 | 3 |
| A10 | 1 | 1 | 1 | 3 | 1 | 1 | 1 | 3 | 1 | 0 | 0 | 1 |
| A11 | 0 | 1 | 1 | 2 | 1 | 1 | 1 | 3 | 0 | 0 | 0 | 0 |
| A12 | 1 | 1 | 1 | 3 | 1 | 0 | 1 | 2 | 1 | 1 | 1 | 3 |
| A13 | 0 | 0 | 0 | 0 | 0 | 1 | 1 | 2 | 0 | 0 | 0 | 0 |
| A14 | 0 | 0 | 0 | 0 | 1 | 1 | 1 | 3 | 0 | 0 | 0 | 0 |
| A15 | 0 | 1 | 0 | 1 | 1 | 1 | 1 | 3 | 1 | 1 | 1 | 3 |
| **SUM** | 8 | 8 | 8 | **24** | 11 | 13 | 12 | **36** | 7 | 6 | 5 | **18** |
| **Percentage** | 53.3% | 53.3% | 53.3% | **53.3%** | 73.3% | 86.7% | 80.0% | **80.0%** | 46.7% | 40.0% | 33.3% | **40.0%** |

[*] 1 (green color) means "Correct", 0 (red color) means "Not correct".

In these "original" questions of Set A, GPT-4 came in the first place as it gave 36 correct answers out of 45 attempts (80.0% success rate), while GPT-3.5 came second, managing to give 24 correct answers (53.3%) and Bard gave 18 correct answers (40.0%). Only in one question (6.7%), A09, all models were correct in all their three attempts. GPT-4 got "all correct" (3 out of 3 attempts) in 10 out of 15 questions (66.7%). Both ChatGPT-3.5 and Bard got "all correct" (3 out of 3 attempts) in 5 out of 15 questions (33.3%).

Similarly, Table 3 presents the scores of each chatbot in the second set of questions (Set B, 15 "Published" problems), for each question, and the relevant sums. In Set B, Bard came first giving 36 correct answers out of 45 attempts (80% success rate), while GPT-4 came second, managing to give 28 correct answers (62.2%) and ChatGPT-3.5 gave 19 correct answers (42.2%). The success rate of Bard was impressive in these problems. Only in 2 questions (13.3%), B05 and B14, all models were correct in all their three attempts, while Bard got "all correct" (3 out of 3 attempts) in 9 out of 15 questions (60%). Similarly, ChatGPT-3.5 got "all correct" in 3 questions (20%) and GPT-4 got "all correct" in 8 out of 15 questions (53.3%).

**Table 3**. Responses of the three chatbots to the questions of Set B ("Published" problems) [*].

| Question | Chat GPT-3.5 | | | | GPT-4 | | | | Bard | | | |
|---|---|---|---|---|---|---|---|---|---|---|---|---|
| | #1 | #2 | #3 | SUM | #1 | #2 | #3 | SUM | #1 | #2 | #3 | SUM |
| **B01** | 1 | 1 | 0 | 2 | 1 | 1 | 1 | 3 | 0 | 1 | 1 | 2 |
| **B02** | 1 | 0 | 1 | 2 | 1 | 1 | 1 | 3 | 1 | 1 | 1 | 3 |
| **B03** | 0 | 0 | 0 | 0 | 0 | 0 | 0 | 0 | 1 | 0 | 0 | 1 |
| **B04** | 0 | 0 | 1 | 1 | 1 | 0 | 0 | 1 | 1 | 1 | 1 | 3 |
| **B05** | 1 | 1 | 1 | 3 | 1 | 1 | 1 | 3 | 1 | 1 | 1 | 3 |
| **B06** | 0 | 0 | 0 | 0 | 1 | 0 | 1 | 2 | 1 | 1 | 1 | 3 |
| **B07** | 0 | 0 | 0 | 0 | 1 | 1 | 1 | 3 | 0 | 0 | 1 | 1 |
| **B08** | 0 | 1 | 0 | 1 | 0 | 0 | 1 | 1 | 1 | 1 | 1 | 3 |
| **B09** | 0 | 0 | 0 | 0 | 0 | 0 | 0 | 0 | 1 | 1 | 1 | 3 |
| **B10** | 0 | 0 | 0 | 0 | 0 | 0 | 0 | 0 | 1 | 1 | 1 | 3 |
| **B11** | 1 | 1 | 1 | 3 | 1 | 1 | 1 | 3 | 1 | 0 | 1 | 2 |
| **B12** | 0 | 1 | 1 | 2 | 1 | 1 | 1 | 3 | 1 | 1 | 0 | 2 |
| **B13** | 1 | 1 | 0 | 2 | 1 | 1 | 1 | 3 | 0 | 0 | 1 | 1 |
| **B14** | 1 | 1 | 1 | 3 | 1 | 1 | 1 | 3 | 1 | 1 | 1 | 3 |
| **B15** | 0 | 0 | 0 | 0 | 0 | 0 | 0 | 0 | 1 | 1 | 1 | 3 |
| **SUM** | 6 | 7 | 6 | **19** | 10 | 8 | 10 | **28** | 12 | 11 | 13 | **36** |
| **Percentage** | 40.0% | 46.7% | 40.0% | **42.2%** | 66.7% | 53.3% | 66.7% | **62.2%** | 80.0% | 73.3% | 86.7% | **80.0%** |

[*] 1 (green color) means "Correct", 0 (red color) means "Not correct".

Figure 1 presents an illustration of the same results as the number of correct answers each chatbot gave for every set, in each of the three rounds (left column), and overall (right column). Comparing the performance of the chatbots in the questions of Set B with Set A, we see that ChatGPT-3.5 fell from 24 correct answers in Set A to 19 correct answers in Set B (20.8% decrease) and similarly the performance of ChatGPT-4 fell from 35 correct answers to 28 (20% decrease), which shows a consistency and that the problems of Set B were probably harder than the ones of Set A. Impressively, the performance of Bard went up from 18 correct answers for the "Original" Set A questions to 36 for the "Published" Set B questions, a remarkable improvement, despite the probable increased difficulty of these problems.

It is obvious that Bard is much better in handling questions that have been already published and answered online, than original questions that have not yet been published. The same is not true for the other two models, which showed a worse performance in the "Published" questions in comparison to the "Original" questions. This is probably because (i) the published questions were in fact harder than the original questions, and (ii) the two GPT chatbots do not have direct access to the internet, in contrast to Bard which does. Indeed, there is an important difference between Bard and the ChatGPT chatbots. Bard can access Google's search engine, while ChatGPT (both versions) has no internet access and has only been trained on information available up to 2021.

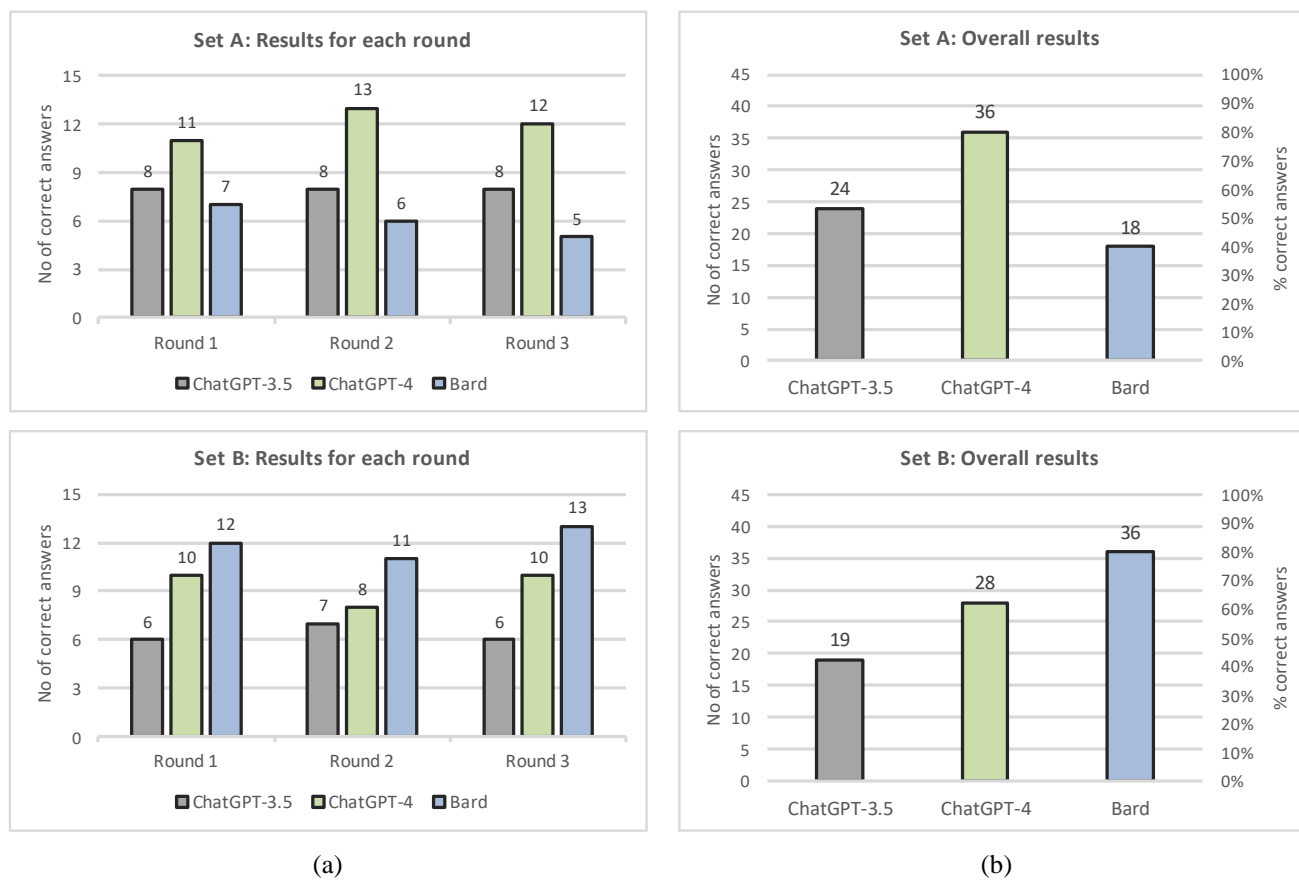

(a)                                                                                                 (b)

**Figure 1**. Performance of each chatbot for the two sets of questions
(a) Results for each of the three rounds, (b) Overall results (all rounds).

## 6    Conclusions

In this study, we compared the performance of three leading chatbot systems, namely ChatGPT-3.5, ChatGPT-4, and Google Bard, in solving math and logic problems. Our aim was to evaluate their understanding, problem-solving abilities, and overall effectiveness in tackling a diverse range of problems. Our findings revealed that all three chatbots demonstrated to some extent an ability to understand and process math and logic problems, with some exceptions and limitations. These models can be used to solve basic mathematics and logic problems, as they have learned to perform simple calculations and understand logical concepts from their training data. However, their capabilities in this domain have certain limitations. For straightforward arithmetic, algebraic expressions, or basic logic puzzles, they may provide accurate solutions, again not every single time. For more complex mathematical problems or advanced logic tasks, their performance may not be as reliable.

ChatGPT-4 clearly outperformed ChatGPT-3.5 in terms of accuracy and handling complex problems. The improved performance of ChatGPT-4 can be attributed to its larger model size, enhanced context understanding, and better fine-tuning compared to its predecessor. Nevertheless, ChatGPT-3.5 is much faster than ChatGPT-4, which is rather slow in generating its responses, probably again due to its larger model size. Bard on the other hand is fast, as it generates three responses at once, but it showed the poorest performance in the set of the original problems, although it exhibited the best performance in the set of published problems which can be found on the internet. Bard has direct access to the internet and to Google search which gives it a competitive advantage when it comes to problems that have been published online, together with their solution.

Despite the recent advancements in chatbot technology, certain limitations were observed in all three chatbots. Complex mathematical problems and those requiring advanced logical reasoning still posed challenges. Even some simple problems appeared to be challenging or hard for the chatbots. Moreover, occasional errors in understanding the problem or misinterpreting user input were observed, highlighting the need for further improvement in natural language understanding. It is essential to understand that these chatbots are primarily language models, not specialized tools for mathematics or logic. While they can demonstrate some problem-solving abilities in these areas, dedicated software or specialized models would be better suited for more complex or advanced tasks in mathematics and logic.

Another problem we observed is that in many cases, the solution the chatbots provide is very long, detailed, and written in a "professional" way, but it still may be completely wrong, or make no sense at all when examined more carefully. This may fool a human to think that such a detailed and long solution would be correct, so extra caution is needed when we use such tools for solving similar exercises. In other words, a chatbot will rarely claim that it does not know the answer to a problem, and will not state its confidence in its solution, like a human would normally do. It will simply give an answer, and the user is not able to know whether this answer can be considered trustworthy or not.

Another issue has to do with the consistency of the responses. In many cases, a model would correctly respond one time but would miserably fail in the very next attempt. There is no warranty that in a given attempt the model will get it correctly. This is a problem especially for questions where we do not know the exact answer and we would rely on the response of the chatbot to get it.

Overall, our study demonstrated the progress made in the field of AI chatbots, with the three chatbots showcasing notable advancements in reasoning and solving math and logic problems. However, there remains room for improvement in terms of accuracy, handling complex problems, and natural language understanding. Future research should focus on addressing these limitations and exploring ways to enhance the chatbots' learning capabilities, enabling them to better adapt to users' needs and provide more accurate, efficient, and reliable assistance in solving math and logic problems.

## List of abbreviations

The following table describes the meaning of various abbreviations and acronyms used throughout the paper.

| Abbreviation | Definition |
|---|---|
| AI | Artificial intelligence |
| DL | Deep learning |
| GPT | Generative pre-trained transformer |
| LLM | Large language model |
| ML | Machine learning |
| NLP | Natural language processing |

**Conflict of Interest**: The authors declare that the research was conducted in the absence of any commercial or financial relationships that could be construed as a potential conflict of interest.

**Data Availability Statement**: The dataset containing the 30 questions, correct answers, explanations and the full answers of all chatbots in all questions can be found in [14].